\pdfoutput=1

\documentclass[11pt]{article}

\usepackage{EMNLP2022}

\usepackage{times}
\usepackage{latexsym}

\usepackage[T1]{fontenc}

\usepackage[utf8]{inputenc}

\usepackage{microtype}

\usepackage{inconsolata}

\usepackage{amsmath}
\usepackage{booktabs}
\usepackage{multirow}

\usepackage{booktabs}
\usepackage{graphicx}
\usepackage{xspace}
\usepackage{makecell}
\usepackage[normalem]{ulem}
\usepackage{tabularx}
\newcommand\setrow[1]{\gdef\rowmac{#1}#1\ignorespaces}
\newcommand\clearrow{\global\let\rowmac\relax}
\clearrow

\newcommand{\sr}[3]{$#1\!#2\!#3$}   

\definecolor{forestgreen}{RGB}{34, 139, 34}
\definecolor{olive}{RGB}{128, 128, 0}
\definecolor{deepskyblue}{RGB}{0, 191, 255}
\definecolor{bbrown}{RGB}{165, 42, 42}

\title{Zero-Shot Text Classification with Self-Training}

\author{\bf{Ariel Gera\thanks{\ \ These authors contributed equally to this work.}, Alon Halfon\footnotemark[1], Eyal Shnarch\footnotemark[1], Yotam Perlitz\footnotemark[1],} \\
\bf{Liat Ein-Dor, Noam Slonim} \\
\\
IBM Research \\
\{ariel.gera1, yotam.perlitz\}@ibm.com,
\\\{alonhal, eyals, liate, noams\}@il.ibm.com}

\begin{document}
\maketitle
\begin{abstract}
Recent advances in large pretrained language models have increased attention to zero-shot text classification. In particular, models finetuned on natural language inference datasets have been widely adopted as zero-shot classifiers due to their promising results and off-the-shelf availability. However, the fact that such models are unfamiliar with the target task can lead to instability and performance issues.  We propose a plug-and-play method to bridge this gap using a simple self-training approach, requiring only the class names along with an unlabeled dataset, and without the need for domain expertise or trial and error. We show that fine-tuning the zero-shot classifier on its most confident predictions leads to significant performance gains across a wide range of text classification tasks, presumably since self-training adapts the zero-shot model to the task at hand. 
\end{abstract}

\section{Introduction} \label{sec:bg}
Large language models have revolutionized the field of natural language processing, leading to great leaps in performance across the NLP task landscape \citep{BERT, raffel2020exploring, brown2020language}.
The pretrain-finetune paradigm has led to a significant reduction in the amount of labeled data required for obtaining high performance on downstream tasks. However, the need to collect labeled examples for each target task remains an obstacle, limiting the 
usage of language models in practice, at scale. 

Thus, the more ambitious vision of a \textit{general-purpose zero-shot model} -- one that can tackle many different tasks without requiring labeled data -- has become an enticing goal for the community. This notion is increasingly gaining attention, 
with recent works suggesting new paradigms that aim to utilize the language understanding capabilities of large models for the zero-shot scenario.

In their pioneering work on more general-purpose zero-shot models,
\citet{yin-etal-2019-benchmarking} propose to formulate text classification tasks as a textual entailment problem \cite{dagan2005pascal}. This mapping enables using a model trained on natural language inference (NLI) as a zero-shot text classifier for a wide variety of unseen downstream tasks. 
The underlying idea is fairly intuitive. To determine if a particular text should be assigned to, e.g., the "sports" class or the "politics" class, one constructs sentences such as "This text is about sports" and "This text is about politics", respectively;
the model prediction as to which one is most entailed by the original text 
can then be used
to determine the predicted class label.
Similarly, some recent works have tried to map even more varied types of NLP tasks into a unified cross-task format \citep{wei2022finetuned, zhong-etal-2021-adapting-language, bragg2021flex, sanh2022multitask}. Such unified task formats enable ``meta-tuning'' a model using existing labeled data from different tasks. By teaching the model to solve the broader ``meta-task'', it is then able to cope with a wide variety of unseen tasks at inference time.

While zero-shot models hold great promise by eliminating the burden of collecting task-specific labeled data, they often still come at the cost of providing mediocre performance compared to models trained in the conventional supervised learning paradigm. Thus, 
improving the prediction performance of zero-shot models is of great practical importance. 
One of the simplest and most effective approaches for improving performance of classifiers is \textit{self-training} \citep{scudder1965probability}. In this paradigm, a model's own predictions on unlabelled data are leveraged for creating pseudo-labels, which are then used for further training the model.

In the original setting of self-training, some labeled data is available for training an initial classifier, and the predictions of the classifier on unlabeled data are used for data augmentation \citep{van2020survey}. More recently, the use of self-training has been extended to the scenario of unsupervised domain adaptation, where labeled data is available only for a source domain, and only unlabeled data is available for the target domain (e.g., \citealp{du-etal-2021-self, zou2019confidence}). 

Here, we aim to study self-training as a method for improving general-purpose zero-shot models, by adapting them to the task at hand. Given the distinct properties of such models, applying self-training in this scenario is not trivial and poses unique challenges. Our approach can be viewed as a further extension of self-training -- from unsupervised domain-adaptation to unsupervised task-adaptation, where only unlabeled data is available for the target task.

As prominent representatives of general-purpose zero-shot models, in this work we focus on NLI-based models \citep{yin-etal-2019-benchmarking}, which are increasingly being utilized for zero-shot classification \citep{davison-HF-blog, sainz-rigau-2021-ask2transformers, basile-etal-2021-probabilistic}. 
To the best of our knowledge, this is the first work that explores self-training in the context of general-purpose zero-shot models. We release our code\footnote{\url{https://github.com/IBM/zero-shot-classification-boost-with-self-training}}, including access to all datasets, and an associated automatic evaluation framework, aiming to facilitate further research along the lines explored here.

\begin{figure}[t]
\begin{center}
\includegraphics[width=1\columnwidth]{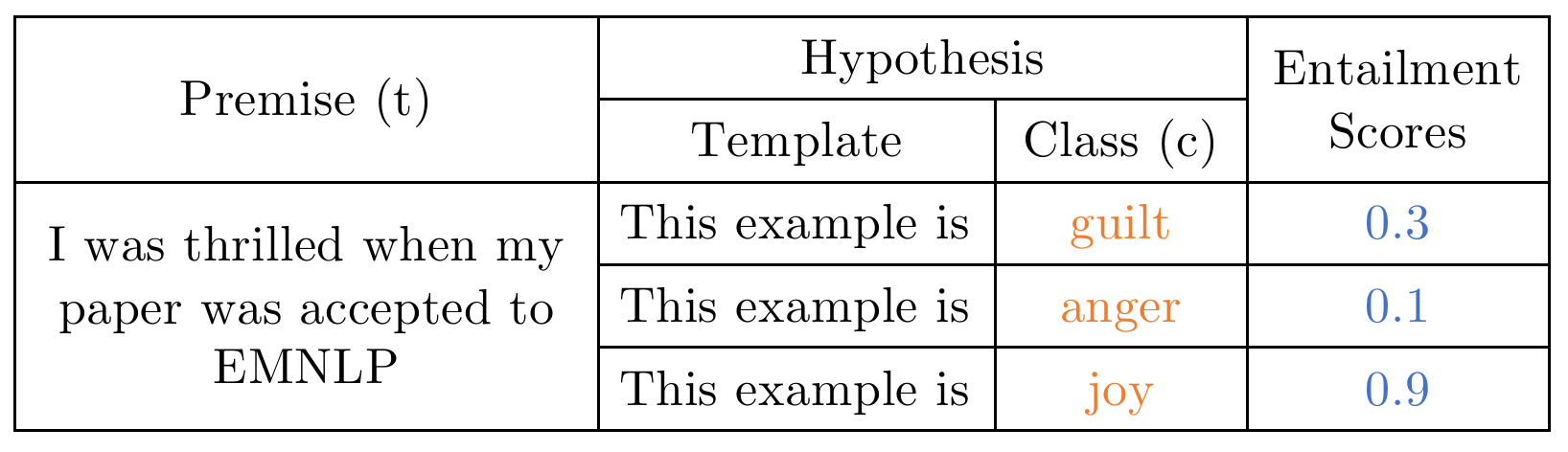}
\caption{\textbf{Entailment-based zero-shot classification.} A text $t$ serves as the Premise while the Hypothesis is created by placing the class name $c$ within a template. The class with the highest entailment score, \textit{joy} in this example, is taken as the label for $t$.}
\label{fig:zs entailment}
\end{center}
\end{figure}

\section{Self-Training of Zero-Shot Text Classifiers}
\label{sec:method}

In self-training, a model $M$ is applied on a collection of unlabeled examples $U$. The instances on which $M$ provides the most confident predictions are taken as a pseudo-labeled training set. This set is used to re-train $M$, giving rise to a new model, $M'$. This procedure can be repeated to obtain $M''$, and so on, in an iterative manner. 

Next, we describe the motivation of applying self-training to zero-shot text classifiers, and the details of our approach. 
\subsection{Motivation} \label{ssec:motivation}
We hypothesize that self-training brings forth unique benefits to general-purpose zero-shot models, going beyond data augmentation and the exposure to the \emph{target domain}.

A zero-shot model, as implied by its name, has never been directly exposed to the task it should perform.
Moreover, one should expect significant differences between the characteristics and distributions of the task(s) used to create the general-purpose model, and those of the 
downstream task.
Self-training may help bridge this gap, by adapting the model to the properties of the 
\emph{target task}.

Specifically, for NLI-based classification models \citep{yin-etal-2019-benchmarking}, which are at the focus of this work, self-training may provide two important benefits, discussed next. 
\paragraph{} \textbf{Exposure to class names.} As a language model, the zero-shot model presumably embodies some knowledge about the meaning of the target class names, considering each class name independently; however, chances are it has never been trained to consider their potential interactions. Pseudo-labeled examples, obtained via self-training, can force the zero-shot model to contrast the class names with one another, and to learn more subtle distinctions that will be required at test time. As a simple example, 'guilt' and 'shame' may often be considered synonyms, but represent two distinct classes in one of our datasets. Explicit exposure to even weakly labeled data is presumably essential to learn such distinctions. 
\paragraph{}\textbf{Exposure to the task and template/prompt.}
Entailment-based models are originally trained on general NLI datasets, which aim to capture a broad and diverse range of textual entailment instances. Utilizing these models for text classification implies that they should focus on a narrower set of entailment types, namely those that map to 
the text classification problem under consideration.
Moreover, the application of these models as zero-shot classifiers involves the use of generic hypothesis templates that aim to formulate the downstream classification task in terms of textual entailment 
-- e.g., "This text is about X".
Both the relevant entailment sub-types, and the generic templates used at test time, are presumably not common in the data used to train the model.
Thus, self-training exposes the model to the specific hypothesis template that will be used for text classification, as well as to the underlying distribution of 
text classification entailment problems it will need to face.

\subsection{Our approach} \label{ssec:approach}
We consider an entailment-based zero-shot model, $M$, for a multi-class classification task, with a set of target class names $C$.

\citet{yin-etal-2019-benchmarking} proposed to map the text classification task into an entailment task, as depicted in Fig. \ref{fig:zs entailment}. Specifically, a target text, $t$, is taken as the premise. For every class \sr{c}{\in}{C}, a hypothesis is constructed from a template such as ``\texttt{This~example~is~$c$}'' (e.g., ``This example is \textit{joy}'', 
or ``This example is \textit{anger}''). The entailment model is presented with $t$ and a set of hypotheses that correspond to the different classes. The class whose hypothesis receives the top entailment score is predicted as the label for $t$ (see Fig. \ref{fig:zs entailment}).

We further assume a collection of unlabeled examples $U$ is available. Following the entailment approach, we generate pseudo-labeled examples from $U$ based on the predictions given by $M$. 

First, for each \sr{u}{\in}{U} and each class name \sr{c}{\in}{C} we obtain $S_{uc}$, the confidence score for $u$ entailing the hypothesis constructed for $c$ (entailment score in Fig. \ref{fig:zs entailment}). In other words, $S_{uc}$ represents the confidence of assigning $u$ to $c$.

\subsubsection{Selecting positive examples}
Our goal is to collect for each class $c$, a set of $n$ pseudo-labeled positive examples in $U$. As common in self-training, we aim to focus on the most confident predictions. We follow a ``Best-versus-Second-Best'' approach, as in \citet{slonim2011active}. 

To that end, we first consider all  examples in $U$ for which $c$ obtained the top entailment score, i.e., $S_{uc} > S_{u{c'}}, \forall {c' \neq c}$. Next, we focus our attention on examples that maximize the delta between the top ranked class and the second highest class (in Fig.~\ref{fig:zs entailment}, the delta is between the entailment score for \textit{joy} and \textit{guilt}). Loosely speaking, such examples correspond to points farthest from the decision boundary, and thus the points that the model is most certain about. Assuming $c$ and $c'$ are the top-ranked and second-ranked classes for $u$, respectively, let $\delta_{uc} = S_{uc} - S_{uc'}$. 

Next, for a given class $c$, we sort all examples in $U$ for which $c$ was the top-ranked class by $\delta_{uc}$ in a decreasing order, and select the top $n$ examples as the positive examples for class $c$.

In order to utilize these pseudo-labeled examples as training examples for the 
entailment model $M$, we use a similar transformation to the one described above -- the example is taken as the premise, the class name is incorporated into the hypothesis template, and the premise-hypothesis pair is assigned the \textit{entail} pseudo-label.

\subsubsection{Selecting negative examples} \label{ssec:neg-examples}
To train the entailment model to contrast between classes we need to generate negative entailment examples, with the \textit{contradict} pseudo-label. For that, we examine four approaches:
\begin{description}
    \item [Contrast-random] For each \textit{entail} pair for a hypothesis based on $c$, add a pair with the \textit{contradict} label, which is composed of the same premise, and a hypothesis in which $c$ is replaced \textbf{at random} with another class.
    \item [Contrast-closest] For each \textit{entail} pair 
    for a hypothesis based on $c$,
    add a pair with the \textit{contradict} label, which is composed of the same premise, and a hypothesis in which $c$ is replaced with the class receiving the \textbf{second-highest} entailment score for this premise (\textit{guilt} in the example of Fig. \ref{fig:zs entailment}).
    \item [Contrast-furthest] For each \textit{entail} pair 
    for a hypothesis based on $c$, 
    add a pair with the \textit{contradict} label, which is composed of the same premise, and a hypothesis in which $c$ is replaced with the class receiving the \textbf{lowest} entailment score for this premise (\textit{anger} in the example of Fig. \ref{fig:zs entailment}).
    \item [Contrast-all] For each \textit{entail} pair 
    for a hypothesis based on $c$, 
    add $\vert C \vert -1$ pairs with the \textit{contradict} label, all with same premise and a hypothesis in which $c$ is replaced with each of the other target class \sr{c'}{\ne}{c}. Note that for datasets with a large number of classes, this setting significantly increases the size of the training set, and correspondingly the run time.
\end{description}
The full training data, including both \textit{entail} and \textit{contradict} pseudo-labeled examples, is used to fine-tune the general entailment model $M$, yielding an entailment zero-shot model $M'$ that has been adapted to the target task. We continue this procedure in iterations: we generate a new pseudo-labeled dataset based on the predictions of $M'$, which is then fine-tuned to generate $M''$, and so forth.

\subsubsection{Balancing noise and informativeness with token masking} \label{ssec:informativeness_vs_noise}

Self-training relies on a delicate balance. On the one hand, the pseudo-labels are noisy. Training on noisy data may lead to overconfidence and propagation of errors \citep{zou2019confidence}. Therefore, a standard self-training practice is to take the most confident predictions, which are presumed to be less noisy. 
On the other hand, the most confident examples are more likely to be the easy and less informative ones, and thus less useful for training \citep{hajmohammadi2015combination, mukherjee-et-al-2020-uncertainty}.

With zero-shot models, this trade-off becomes even more pronounced. As these models were not trained on the target task, the pseudo-labels that are obtained from their predictions are likely to be noisier than those obtained from a model trained on some labeled data.
Thus, with zero-shot models we are compelled to raise the confidence bar in order to obtain 
pseudo-labels of reasonable quality, which in turn may focus the training on the easy and thus less informative examples. 

To increase the informativeness of the selected examples, we apply the following heuristic: in each example we identify the token which is the most similar to the 
positive class name assigned to this example, and mask it. In the example of Fig. \ref{fig:zs entailment}, the word \textit{thrilled} will be 
masked when this example is used as a positive or as a negative example for the class "joy". 
By masking those most similar tokens, the selected examples become more challenging, 
and the model is forced to rely on other signals -- e.g., in Fig. \ref{fig:zs entailment}, on the understanding that the event of a paper getting accepted to a conference is a joyful one.

\section{Experimental Setup} \label{sec:experiment}
\subsection{Datasets and Tasks} \label{subsec:datasets and tasks}
We experiment with 8 datasets representing a variety of text classification tasks: 20 newsgroup \citep{Lang95newsgroup}, AG's news \citep{ds-ag-news:15}, Amazon reviews \citep{mcauley2013amazon}, DBPedia \citep{ds-ag-news:15}, GoEmotions \citep{demszky-etal-2020-goemotions}, IMDB reviews \citep{maas-etal-2011-imdb}, ISEAR \citep{ds-isear:15}, and Yahoo! Answers \citep{ds-ag-news:15}. All datasets, except GoEmotions, are balanced.
Generally, the original dataset class names were used to describe the target labels for zero-shot inference\footnote{Only where labels could not be used as-is (e.g., \textit{soc.religion.christian} in 20 newsgroup) we manually picked a more readable name to represent the class.}. We report results on the test set of each dataset (the labels of the train sets were not used as there is no training in our method); preliminary experiments were conducted on separate development sets.
Since we aim for a practical setting with lower computational costs, we limit the size of our unlabeled set $U$ to a maximum of 10K examples sampled from the full training set of each dataset. 
For details on the dataset sizes, task types, and label names, see App.~\ref{app:datasets}. 

\subsection{Zero-Shot Models}
We evaluate $3$ off-the-shelf entailment models, trained on the MNLI \citep{williams-etal-2018-mnli} dataset: \textit{roberta-large-mnli}, \textit{deberta-large-mnli-zero-cls}, and \textit{bart-large-mnli}. To infer zero-shot predictions from these models with respect to the target labels we rely on the dedicated zero-shot classification pipeline from the Hugging Face Transformers library\footnote{\url{https://huggingface.co/zero-shot/}}, using the default hypothesis template \textit{"This example is []."}.

\subsection{Implementation Details}
Each experiment is repeated $5$ times, with each repetition using a different random seed.
All models are fine-tuned for one epoch with a learning rate of $2\times10^{-5}$ and a batch size of $32$, using the AdamW optimizer \cite{kingma2014adam} and cross entropy loss to optimize the models.
A single NVIDIA A100 GPU was used for fine-tuning and inference. We base our implementation on Hugging Face Transformers \cite{wolf2019huggingface} version 4.16.2 and pytorch \cite{paszke2019pytorch} version 1.10. 

\subsection{Token masking} \label{ssec:masking}

\begin{figure}[t]
\begin{center}
\includegraphics[width=1\columnwidth]{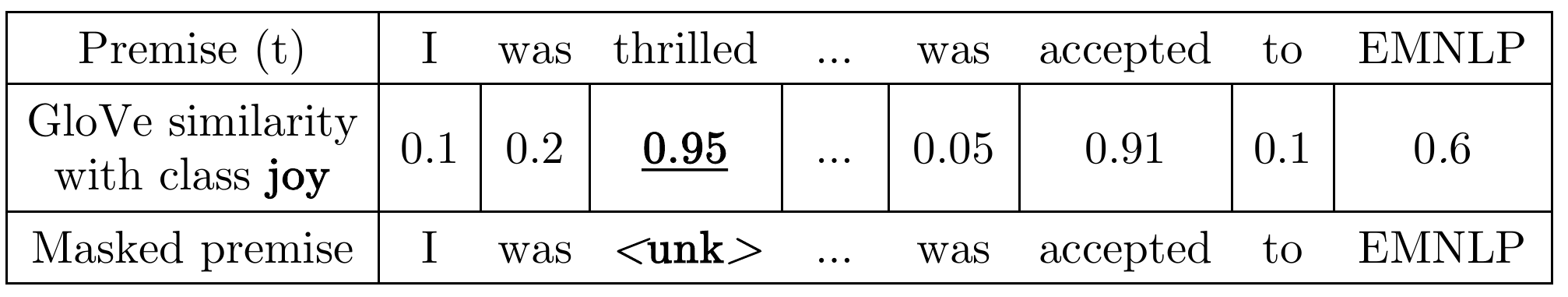}
\caption{\textbf{GloVe token masking.} All words of the Premise $t$ are scored according to the GloVe similarity with class name $c$. The word with the top similarity score is replaced with an <UNK> token.}
\label{fig:zs glove masking}
\end{center}
\end{figure}

\begin{table*}[h]
\centering
\begin{tabular}{|l|>{\rowmac}c>{\rowmac}c>{\rowmac}c>{\rowmac}c>{\rowmac}c>{\rowmac}c>{\rowmac}c>{\rowmac}c|>{\rowmac}c<{\clearrow}|}
\hline
 & 20NG & AG & DBPed. & Yahoo & GoEmo. & ISEAR & Amazon & IMDB & Avg.
\\
\hline
BART & 45.0 & 66.2 & 74.7 & 48.3 & 19.7 & 56.0 & 93.3 & 91.7 & 61.9  \\
+Self-training & \setrow{\bfseries} 63.7 & 74.2 & 94.1 & 61.0 & 28.1 & 65.3 & 94.7 & 92.2 & 71.7  \\
\hline
DeBERTa & 50.8 & 73.2 & 74.7 & 43.5 & 25.0 & 58.5 & 92.2 & 90.3 & 63.5  \\
+Self-training & \setrow{\bfseries} 67.8 & 81.4 & 94.5 & 62.0 & 30.3 & 59.5 & 95.0 & 92.3 & 72.8  \\
\hline
RoBERTa & 34.1 & 62.4 & 69.8 & 35.9 & 21.4 & 52.0 & 93.1 & 90.7 & 57.4  \\
+Self-training & \setrow{\bfseries} 65.8 & 76.5 & 92.2 & 59.8 & 29.3 & 56.7 & 94.3 & 92.5 & 70.9  \\

\hline
\end{tabular}
\caption{\textbf{Zero-shot classification accuracy of entailment models.} For each zero-shot entailment model and dataset, we compare the test accuracy of the off-the-shelf model to its accuracy after $2$ iterations of self-training. RoBERTa, DeBERTa, and BART correspond to the following models from Hugging Face Hub: \textit{roberta-large-mnli}, \textit{deberta-large-mnli-zero-cls}, and \textit{bart-large-mnli}. Each result is the average of $5$ repetitions using different random seeds.
\label{tab:all_res}}
\end{table*}

As mentioned in \ref{ssec:informativeness_vs_noise}, when collecting pseudo-labeled examples from the model predictions, we mask a token in the example texts based on similarity to the predicted class names. For each example we extract the GloVe \citep{Pennington2014GloveGV} representations for each token in the example text, and for the predicted class name. Where the class name is an ngram, we average over its unigrams.
Representations are extracted using the \textit{en-core-web-lg} model from the spacy library, after removing punctuation and stopwords.

As illustrated in Fig.~\ref{fig:zs glove masking}, for each example, we select the token with the largest GloVe similarity to the class name. This token is then masked from the text by replacing it with the model's special unknown token (\textit{<unk>} for RoBERTa and BART, \textit{[UNK]} for DeBERTa).

\section{Experimental Results} \label{sec:results}

We set $n$, the number of training examples per class, to be $1\%$ of the unlabeled set $U$ (i.e., \sr{n}{=}{100} for a $U$ of size $10k$). For each dataset and zero-shot model, we perform two iterations of self-training.\footnote{We did not observe consistent improvements after the second iteration.}  
We test $4$ settings of adding \textit{contradict} examples as described in 
Section \ref{ssec:neg-examples}.

Classification accuracy before and after self-training for all models using the \textit{Contrast-random} setting is shown in Table \ref{tab:all_res}. The results demonstrate a clear and significant benefit to the self-training process, across all models and datasets. Significance was tested with paired t-tests to compare accuracy with and without self-training, pooling together all datasets and seeds for each of the three models.

\begin{figure*}[th]
\begin{center}
\includegraphics[width=1\textwidth]{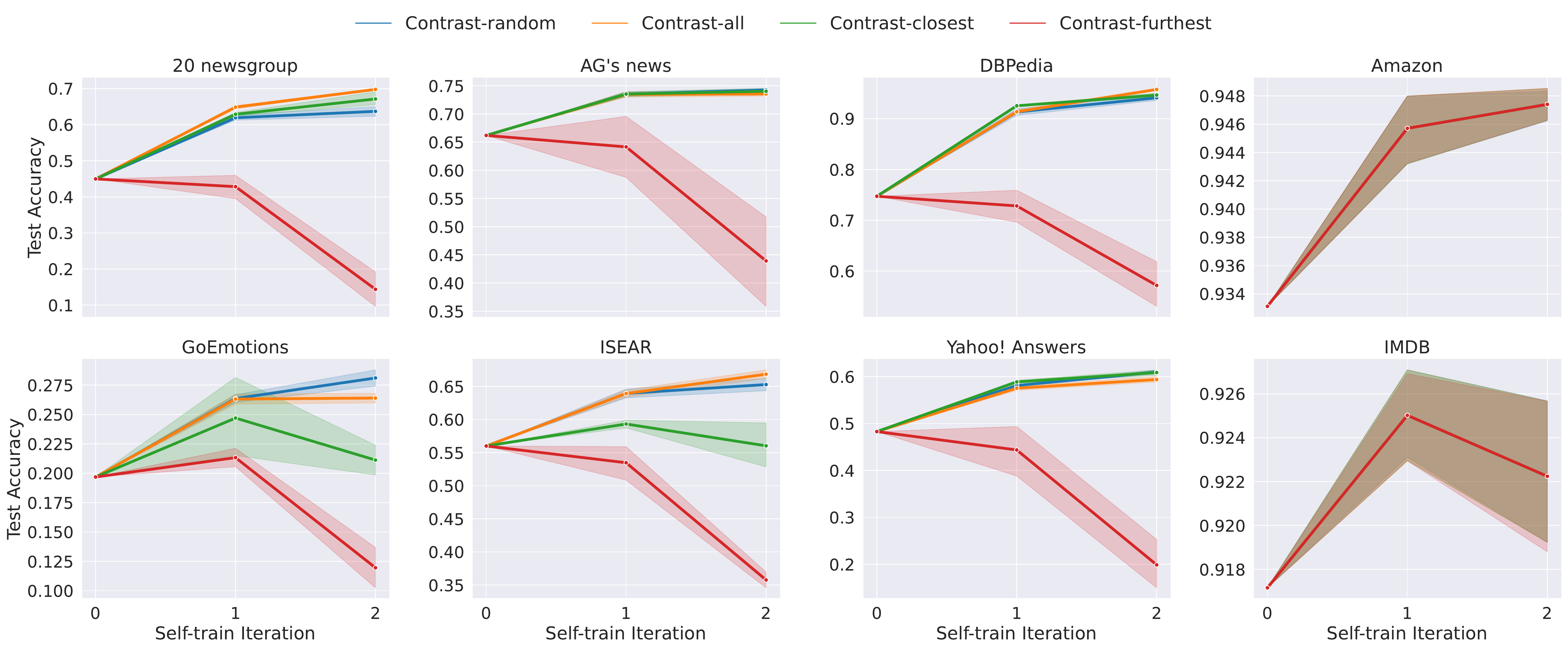}
\caption{\textbf{Self-training with different settings of selecting negative examples.} Lines depict zero-shot classification accuracy ($\pm$SEM, standard error of the mean) of the \textit{bart-large-mnli} model over $2$ self-training iterations, using different approaches for selecting negative examples (see \S\ref{ssec:neg-examples}). Iteration $0$ denotes the accuracy without self-training.}
\label{fig:contrast_method}
\end{center}
\end{figure*}

\begin{figure*}[th]
\begin{center}
\includegraphics[width=0.7\textwidth]{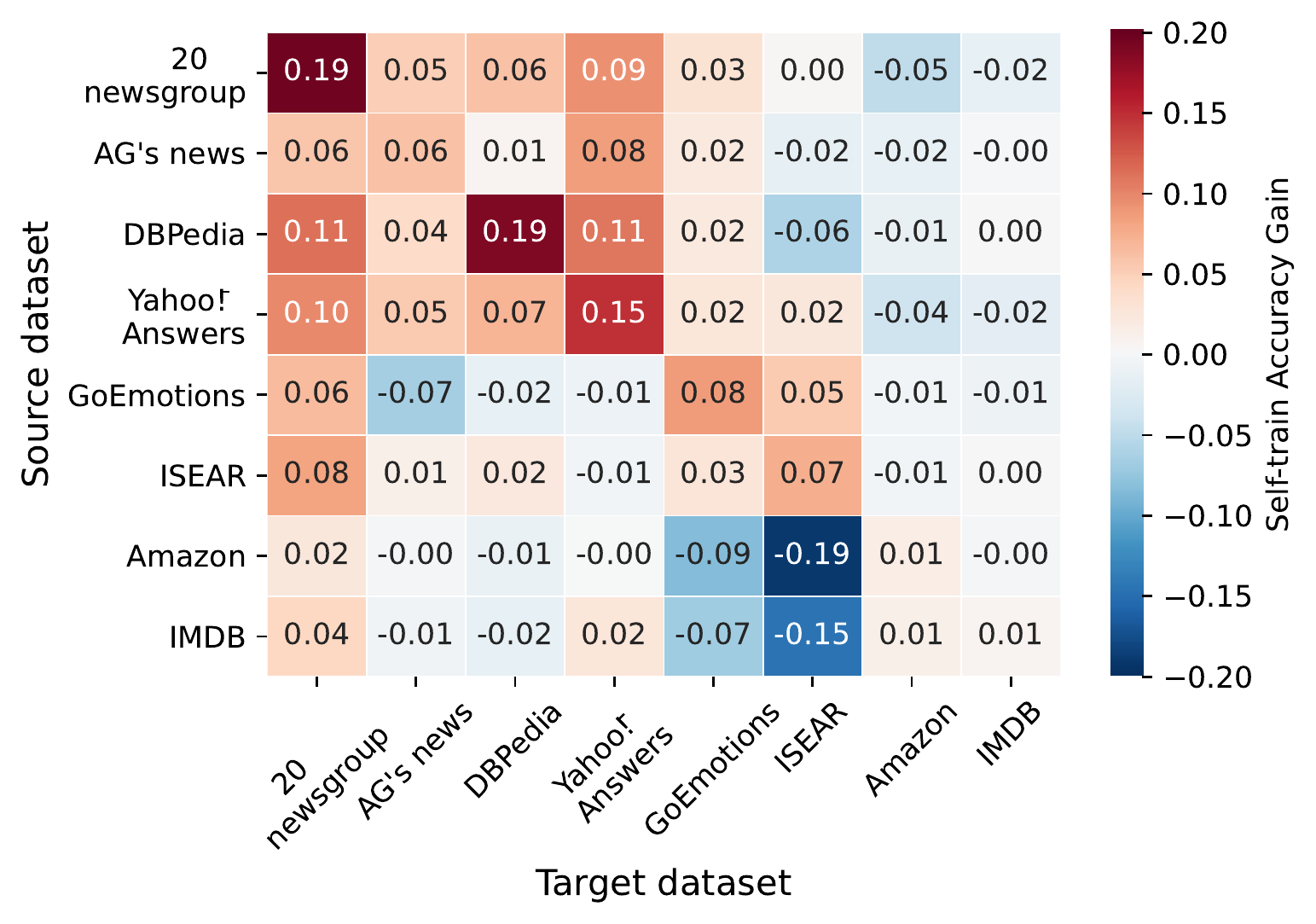}
\caption{\textbf{Cross-task effects of self-training.} The number in each cell is the zero-shot test accuracy over the target dataset after two iterations of self-training over the source dataset, minus the test accuracy over the target dataset with no self-training. The model used is BART (\textit{bart-large-mnli}).
}

\label{fig:cross_evaluation}
\end{center}
\end{figure*}

Fig.~\ref{fig:contrast_method} compares the $4$ settings of selecting negative examples. As can be seen, among the four settings, \textit{Contrast-furthest} yields the poorest results. A possible explanation is that in this setting the negative examples are too trivial for the model. \textit{Contrast-closest} yields better results, probably because the negative examples in this setting are more difficult and thus more informative for the model. However, for this same reason, these pseudo labeled examples are expected to suffer from the largest noise. The best performing settings are \textit{Contrast-all} and \textit{Contrast-random}, which represent a better balance between the informativeness of the negative examples and their noise level. Taking the computational costs into account, \textit{Contrast-random} emerges as the preferred setting.

\begin{table*}[h]
\centering
\begin{tabular}{|l|>{\rowmac}c>{\rowmac}c>{\rowmac}c>{\rowmac}c>{\rowmac}c>{\rowmac}c>{\rowmac}c>{\rowmac}c|>{\rowmac}c<{\clearrow}|}
\hline
 & 20NG & AG & DBPed. & Yahoo & GoEmo. & ISEAR & Amazon & IMDB & Avg.
\\
\hline
BART+ST & 66.7 & 74.3 & 92.8 & 61.1 & 28.9 & 48.0 & 94.6 & 92.2 & 69.8  \\
+Mask & 63.7 & 74.2 & 94.1 & 61.0 & 28.1 & 65.3 & 94.7 & 92.2 & 71.7  \\
\hline
DeBERTa+ST & 70.1 & 79.7 & 93.2 & 62.3 & 30.2 & 52.2 & 94.1 & 92.0 & 71.7  \\
+Mask & 67.8 & 81.4 & 94.5 & 62.0 & 30.3 & 59.5 & 95.0 & 92.3 & 72.8  \\
\hline
RoBERTa+ST & 63.7 & 72.7 & 91.5 & 58.7 & 28.9 & 48.1 & 93.3 & 91.8 & 68.6  \\
+Mask & 65.8 & 76.5 & 92.2 & 59.8 & 29.3 & 56.7 & 94.3 & 92.5 & 70.9  \\

\hline
\end{tabular}
\caption{\textbf{The effect of token masking.} For each zero-shot entailment model and dataset, we compare the test accuracy after $2$ iterations of self-training, with and without applying token masking to the self-training examples. RoBERTa, DeBERTa, and BART correspond to the following models from Hugging Face Hub: \textit{roberta-large-mnli}, \textit{deberta-large-mnli-zero-cls}, and \textit{bart-large-mnli}. Each result is the average of $5$ repetitions using different random seeds.
\label{tab:masking}}
\end{table*}

\section{Analysis} \label{sec:analysis}

\subsection{The contribution of token masking} \label{ssec:contribution_of_masking}
One component of our approach, which aims at increasing the informativeness of pseudo-labeled examples, is the masking of the token closest to the class name (\S\ref{ssec:masking}).

We examine the performance of self-training without the masking procedure. 
A comparison of classification accuracy with and without applying token masking is shown in Table~\ref{tab:masking}. Overall, applying token masking does provide a performance gain as confirmed by a paired t-test ($p=3{\times}{10^{-4}}$, pooling together all models, datasets and seeds). However, as can be seen in Table~\ref{tab:masking}, results do differ across models and datasets. For instance, masking affords more consistent benefits in the case of the RoBERTa entailment model, and has a more pronounced effect in the case of the ISEAR dataset.

The beneficial effect of masking raises the question of whether the pseudo-labeled train set, i.e., the model's most confident predictions, are trivial examples that could just as easily have been obtained using a simple heuristic. To test this, we construct an alternative pseudo-labeled set that is based on a token-level heuristic rather than the model predictions. This example selection method had a substantial negative effect on performance (see App. \ref{app:glove_heuristic} for details).

\subsection{Cross-task effects} \label{ssec:cross-task}
A recurring question in transfer learning is whether fine-tuning on a task $T_i$ can translate to benefits on another task, $T_j$ (e.g., \citealp{phang2018sentence, aghajanyan-etal-2021-muppet}). It is thus interesting to study this question in the present context of self-training zero-shot classifiers. In other words, can exposure to pseudo-labels for task $T_i$ improve zero-shot performance on task $T_j$. One aspect that is specific to our scenario is that fine-tuning with pseudo-labels on $T_i$ exposes the model to the task template which is also used for $T_j$. 

To explore this question, each model that was self-trained on $T_i$ is evaluated over all the other datasets as well. Fig.~\ref{fig:cross_evaluation} depicts the cross-task effects of self-training on performance. 
This analysis reveals that self-training on a different task can be beneficial or harmful. The topical datasets (20 newsgroup, AG's news, DBPedia, and Yahoo! answers) appear to be beneficial to each other; as do the two emotion classification datasets (GoEmotions and ISEAR). In contrast, self-training on sentiment data (Amazon, IMDB) leads to significant degradation in results on the emotion datasets. 
Possibly, this is related to particular characteristics of the reviews domain, along with the sharp binary distinction between positive and negative sentiment, 
as opposed to the subtle nuances that are necessary to distinguish between different types of 
emotions.
\section{Related Work}

Self-training \citep{scudder1965probability} has a long history as a method for semi-supervised learning, where predictions of a supervised classifier on unlabeled examples are used as pseudo-labels to augment the amount of training data. This approach has successfully been applied to a wide range of machine learning problems \citep{van2020survey}. Many variations of self-training have been put forth, varying in terms of the selection strategy of samples to pseudo-label, the amount -- and characteristics -- of the models involved in the procedure, and other specific design decisions \citep{triguero2015self}.

From a more theoretical standpoint, previous works \citep{lee2013pseudo} have described self-training as somewhat equivalent to entropy minimization \citep{grandvalet2004semi}, in that it modifies the model's decision boundaries by driving the model to make more confident predictions.

Aiming for more general-purpose models, that can achieve cross-task generalization and perform in a zero-shot scenario, recent works have proposed different strategies for mapping a range of NLP tasks into a generic and unified framework. \citet{yin-etal-2019-benchmarking} suggest the textual entailment paradigm as one that can encompass different types of text classification tasks. \citet{zhong-etal-2021-adapting-language} map classification tasks to a question-answering format, where each class is formulated as a question and given as a prompt, and the decoder probabilities of the \textit{Yes} and \textit{No} tokens correspond to a positive or negative prediction for the class. They also propose a "meta-tuning" paradigm, where labeled data for different tasks -- formulated in terms of the unified task format -- is utilized in order to teach the model how to solve the generic "meta-task", and thus better cope with unseen tasks at test time. By opting for a generic cross-task format of natural language instructions, \citet{wei2022finetuned} and \citet{sanh2022multitask} extend this notion even further, where meta-tuning on multiple types of NLP tasks enables zero-shot prediction even on tasks of a very different nature from those seen during training.

In the present work we explore the intersection of these two threads -- namely, self-training and general purpose zero-shot models, while focusing on zero-shot text classifiers that use the entailment paradigm, and on a scenario where only unlabeled data is available.

\citet{ye-etal-2020-zero} apply self-training to text classification in order to transfer to unseen classes for which there is no labeled data, and propose a reinforcement learning method for selecting examples to pseudo-label. This scenario differs substantially from ours in that self-training is not applied to an existing general-purpose zero-shot model. In addition, they deal with a setting where labeled data for some of the target classes is available.

Like the present work, \citet{zhou2022prompt} also aim to improve existing general-purpose zero-shot learners by utilizing unlabeled data. Starting from T0, the prompt-based zero-shot learner from \citet{sanh2022multitask}, they use unlabeled texts to apply a \textit{prompt consistency loss}: an example is fed into the model multiple times, each time in the context of a different -- but synonymous -- task prompt; then, the model is trained to assign similar predictions across differently-phrased prompts \citep{zhou2022prompt}. Thus, whereas we explore improving a general-purpose model using a form of self-training, they do so using a variation on the paradigm of consistency training \citep{xie2020unsupervised}.

Some works attempt to improve the performance of general-purpose models within a \textit{few-shot} scenario. For example, \citet{basile-etal-2021-probabilistic} experiment with entailment-based classifiers. They show that compared to a standard pre-trained language model, off-the-shelf entailment models require less labeled 
examples for fine-tuning to reach reasonable performance on an emotion classification task.
\section{Discussion}

In this paper we look at the applicability of self-training for adapting a general-purpose zero-shot model, focusing on the scenario of entailment-based models. We opted for this specific setting due to the high accessibility of these off-the-shelf models. In other words, given that these models are readily available for use, we ask whether self-training provides a straightforward way for practitioners to adapt the general model for their downstream task, using only a modest collection of unlabeled data. We show that in this setting self-training does indeed provide value, delivering significant performance gains for text classification.

The notion of using self-training as a tool to adapt a general-purpose zero-shot model is not specific to entailment models, nor is it limited to classification tasks. Thus, a major avenue for future work would be to explore this combination on models that rely on different kinds of mapping functions or ``meta-tasks'' for formulating downstream tasks within a generic cross-task framework \citep{wei2022finetuned, zhong-etal-2021-adapting-language, bragg2021flex, sanh2022multitask}.

Zero-shot text classification is recently drawing much attention, with prominent recent works showing promising results using different kinds of iterative approaches \citep{meng-etal-2020-text, zhang-etal-2021-weakly}. Such approaches build their zero-shot classifiers from scratch -- and therefore typically require larger unlabeled datasets to perform well -- whereas we aim to utilize and build on the knowledge contained in general-purpose zero-shot classifiers. Exploring ways to combine these differing approaches is left for future work.

Importantly, while our method does assume the existence of a collection of unlabeled examples, our results show that an order of 10K examples is sufficient to benefit from self-training. Moreover, the cross-task effects in section \ref{ssec:cross-task} demonstrate that even unlabeled examples from a similar domain and/or task may be useful in adapting the general-purpose model for the downstream task. Determining the exact conditions under which self-training is useful for adaptation across tasks is a matter for future study. Moreover, it would be interesting to explore the effects of self-training on multiple datasets, akin to works on supervised multi-task fine-tuning (e.g., \citealp{aghajanyan-etal-2021-muppet}).

In our work, we select examples for pseudo-labeling based solely on model confidence 
(see \S\ref{ssec:approach}).
Some self-training works opt for more balanced approaches for 
example selection, aiming for a more diverse and/or more informative set of examples (e.g., \citealp{hajmohammadi2015combination, mukherjee-et-al-2020-uncertainty}). It would be interesting to explore such questions in the zero-shot scenario. In addition, in Section \ref{ssec:approach} we describe our method to select confident examples, namely by looking at the maximal delta between the highest and second highest prediction scores. Other alternatives for choosing confident examples, e.g., by looking at the entropy across classes, could be tested as well.

To conclude, the method we proposed in this paper can boost the performance of entailment-based zero-shot text classifiers, with little effort and a modest amount of domain data. This can prove useful to the many practitioners who benefit from the practicality and accessibility of these models.

\section*{Limitations}

Our focus here is on off-the-shelf models that are highly accessible -- and thus potentially useful -- for practitioners. Nevertheless, these models are quite large, and thus carry a non-negligible computational footprint. For instance, inferring on $10K$ unlabeled samples does require a GPU, limiting access to such approaches and models in practice.

Our work is empirical in nature. As such, we report experimental results, with no theoretical guarantees, and one should recognize the existence of exceptions. In addition, our experimental results are for relatively standard academic benchmarks for text classification. Real-world datasets, especially in specific domains such as legal and healthcare, may pose additional challenges. The practical value of our approach in these cases is yet to be seen. 

We formulate and test our approach in the scenario where each example should be assigned to exactly one class. Applying our method to the multi-label classification scenario might not be straightforward, and may require different ways of selecting examples for the pseudo-labeled set.

Finally, the large scale of our experiments places a non-trivial burden on trying to replicate our results. Moreover, the off-the-shelf models used in our experiments are not guaranteed to be hosted publicly in the future.

\section*{Acknowledgements}
We thank Leshem Choshen for his invaluable insights and feedback as we pursued this research.

\bibliography{custom}
\bibliographystyle{acl_natbib}

\appendix
\section{Datasets} \label{app:datasets}

To complete the details from subsection \ref{subsec:datasets and tasks}, Table~\ref{tab:dataset} shows the statistics of all datasets used while Table~\ref{tab:classnames} details the class names that we used in all our experiments.

\begin{table*}[t]
\centering
\begin{tabular}{*{5}{c}}
\toprule
\textbf{Dataset} & \textbf{Classification Type} & \textbf{\# Classes} & \textbf{\# Unlabeled} & \textbf{\# Test} \\
\midrule
\href{https://docs.google.com/uc?export=download&id=0Bz8a_Dbh9QhbUDNpeUdjb0wxRms}{\textbf{AG News}} & News Topic & 4 & 10,000 & 7,600 \\
\href{https://docs.google.com/uc?export=download&id=0Bz8a_Dbh9QhbQ2Vic1kxMmZZQ1k&confirm=t}{\textbf{DBPedia}} & Wikipedia Topic & 14 & 10,000 & 70,000 \\
\href{https://ai.stanford.edu/~amaas/data/sentiment/}{\textbf{IMDB}} & Movie Review Sentiment & 2 & 10,000 & 25,000 \\
\href{https://jmcauley.ucsd.edu/data/amazon/}{\textbf{Amazon}} & Product Review Sentiment & 2 & 10,000 & 400,000 \\
\href{https://raw.githubusercontent.com/sinmaniphel/py_isear_dataset/master/isear.csv}{\textbf{ISEAR}} & Emotion & 7 & 5366 & 1534 \\
\href{https://github.com/google-research/google-research/tree/master/goemotions}{\textbf{GoEmotions}} & Emotion & 28 &10,000 & 5427 \\
\href{https://scikit-learn.org/0.15/datasets/twenty_newsgroups.html}{\textbf{20 newsgroup}} & News & 20 & 10,000 & 7532 \\
\href{https://docs.google.com/uc?export=download&id=0Bz8a_Dbh9Qhbd2JNdDBsQUdocVU&confirm=t}{\textbf{Yahoo! Answers}} & Question Topic & 10 & 10,000 & 58966 \\

\bottomrule
\end{tabular}
\caption{
Dataset statistics. 
}
\label{tab:dataset}
\end{table*}

\begin{table*}[t]
\centering
\begin{tabular}{|p{0.15\textwidth}|p{0.8\textwidth}|}
\hline
\textbf{Dataset}   & \textbf{Class names used}                                                \\ \hline
ISEAR     & 'anger', 'disgust', 'fear', 'guilt', 'joy', 'sadness', 'shame' \\ \hline
GoEmotions &
  'admiration', 'amusement', 'anger', 'annoyance', 'approval', 'caring', 'confusion', 'curiosity', 'desire', 'disappointment', 'disapproval', 'disgust', 'embarrassment', 'excitement', 'fear', 'gratitude', 'grief', 'joy', 'love', 'nervousness', 'neutral', 'optimism', 'pride', 'realization', 'relief', 'remorse', 'sadness', 'surprise' \\ \hline
AG's news & 'business', 'world', 'sports', 'science and technology'        \\ \hline
Yahoo! Answers &
  'business \& finance', 'computers \& internet', 'education \& reference', 'entertainment \& music', 'family \& relationships', 'health', 'politics \& government', 'science \& mathematics', 'society \& culture', 'sports' \\ \hline
20 newsgroup &
  'atheism', 'computer graphics', 'hockey', 'cryptography', 'electronics', 'medicine', 'space', 'christianity', 'guns', 'middle east', 'politics', 'religion', 'microsoft windows', 'pc hardware', 'mac hardware', 'windows x', 'for sale', 'cars', 'motorcycles', 'baseball' \\ \hline
DBPedia &
  'album', 'animal', 'artist', 'athlete', 'building', 'company', 'educational institution', 'film', 'mean of transportation', 'natural place', 'office holder', 'plant', 'village', 'written work' \\ \hline
Amazon    & 'bad', 'good'                                                  \\ \hline
IMDB      & 'bad', 'good'                                                  \\ \hline
\end{tabular}
\caption{
Dataset class names used in all our experiments. }
\label{tab:classnames}
\end{table*}

\section{Heuristic-based selection} \label{app:glove_heuristic}
\begin{figure*}[th]
\begin{center}
\includegraphics[width=1\textwidth]{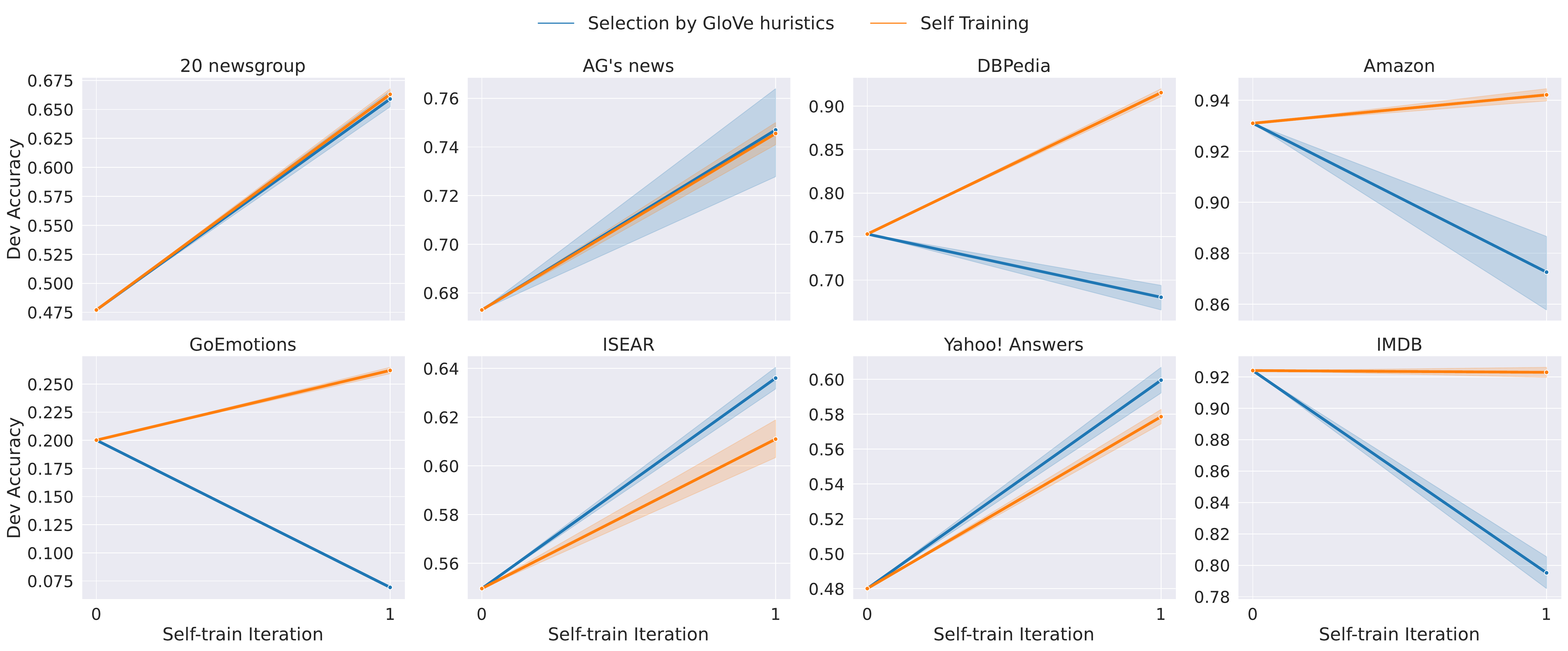}
\caption{\textbf{Self-training versus heuristic-based selection.} Lines depict zero-shot classification accuracy ($\pm$SEM, standard error of the mean) of the \textit{bart-large-mnli} model over a single iteration of fine-tuning over pseudo-labeled examples, using either model top predictions (orange) or a GloVe-based token-level heuristic (blue). Iteration $0$ denotes the accuracy of the off-the-shelf model.}
\label{fig:glove_heuristic}
\end{center}
\end{figure*}

As stated in section \ref{ssec:contribution_of_masking}, we experiment with constructing an alternative pseudo-labeled train set that is based on a token-level heuristic. In this method, the examples are chosen based on GloVe-based similarity to the class names. First, for each example and class we calculate a "GloVe-to-closest-token" score, which is the similarity between the class and the closest token in the example, following a similar protocol as that for finding tokens to mask (cf. \ref{ssec:masking}). Then, for each class $c$ we construct a list of size $n$ of the top candidates: we take the examples where the "GloVe-to-closest-token" score was highest for $c$; these examples are sorted by the difference between the "GloVe-to-closest-token" score for $c$ and for the class with the second highest score,
and the top $n$ examples are selected. We apply masking for the selected examples using the same protocol we use in the self-training approach.
Fig. \ref{fig:glove_heuristic} compares this approach to a single iteration of self-training. As can be seen, for some datasets this pseudo-labeling approach does improve the zero-shot classifier, yet, the results are not consistent across datasets and in $4$ of the datasets applying this approach results in a lower accuracy compared to the zero-shot classifier.

\section{Results on validation set}
Table \ref{tab:dev_results} shows the zero-shot classification accuracy before and after self-training for the validation sets.

\begin{table*}[h]
\centering
\begin{tabular}{|l|>{\rowmac}c>{\rowmac}c>{\rowmac}c>{\rowmac}c>{\rowmac}c>{\rowmac}c>{\rowmac}c>{\rowmac}c|>{\rowmac}c<{\clearrow}|}
\hline
 & 20NG & AG & Amazon & DBPed. & GoEmo. & IMDB & ISEAR & Yahoo & Avg.
\\
\hline
BART & 47.7 & 67.3 & 93.1 & 75.3 & 20.0 & \textbf{92.4} & 55.0 & 48.0 & 62.3  \\
+Self-training & \textbf{65.3} &\textbf{ 75.0 }& \textbf{94.4} & \textbf{93.4 }& \textbf{28.1} & 91.9 & \textbf{61.6} & \textbf{61.5} & \textbf{71.7}  \\
\hline
DeBERTa & 50.7 & 74.5 & 92.2 & 75.1 & 24.6 & 90.6 & 55.2 & 43.0 & 63.2  \\
+Self-training & \setrow{\bfseries} 69.3 & 82.2 & 94.8 & 94.5 & 29.9 & 92.1 & 58.7 & 62.4 & 73.0  \\
\hline
RoBERTa & 32.7 & 63.5 & 92.7 & 69.4 & 21.2 & 90.6 & 49.5 & 35.8 & 56.9  \\
+Self-training & \setrow{\bfseries} 66.7 & 77.0 & 94.2 & 92.1 & 29.1 & 92.9 & 55.2 & 60.3 & 70.9  \\

\hline
\end{tabular}
\caption{\textbf{Zero-shot classification accuracy of entailment models.} For each zero-shot entailment model and dataset, we compare the validation accuracy of the off-the-shelf model to its accuracy after $2$ iterations of self-training. RoBERTa, DeBERTa, and BART correspond to the following models from Hugging Face Hub: \textit{roberta-large-mnli}, \textit{deberta-large-mnli-zero-cls}, and \textit{bart-large-mnli}. Each result is the average of $5$ repetitions using different random seeds.
\label{tab:dev_results}}
\end{table*}

\end{document}